\pdfoutput=1

\documentclass[11pt]{article}

\usepackage{EACL2023}

\usepackage{times}
\usepackage{latexsym}
\usepackage{amsmath}
\usepackage[T1]{fontenc}

\usepackage[utf8]{inputenc}

\usepackage{microtype}

\usepackage{inconsolata}

\usepackage{graphicx}
\usepackage{algorithm}
\usepackage{algorithmic}
\newcommand{\bm}{\mathbf}
\usepackage{booktabs}

\title{NapSS: Paragraph-level Medical Text Simplification via 
Narrative Prompting and Sentence-matching Summarization}

\author{Junru Lu$^1$, Jiazheng Li$^2$, Byron C. Wallace$^3$, Yulan He$^{1,2,4}$ and Gabriele Pergola$^1$ \\
  $^1$Department of Computer Science, University of Warwick, UK\\
  $^2$Department of Informatics, King's College London, UK\\
  $^3$Northeastern University, USA~~~~~$^4$The Alan Turing Institute, UK\\
    \texttt{\{junru.lu, gabriele.pergola\}@warwick.ac.uk} \\
    \texttt{b.wallace@northeastern.edu}, \texttt{\{jiazheng.li, yulan.he\}@kcl.ac.uk}}

\begin{document}
\maketitle

\begin{abstract}
  Accessing medical literature is difficult for laypeople as the content is written for specialists and contains medical jargon.
  Automated text simplification methods offer a potential means to address this issue. 
  In this work, we propose a 
  \emph{summarize-then-simplify} two-stage strategy, which we call \texttt{NapSS}, identifying the relevant content to simplify while ensuring that the original narrative flow is preserved. 
  In this approach, we first generate reference summaries via sentence matching between the original and the simplified abstracts. These summaries are then used to train an extractive summarizer, learning the most relevant content to be simplified.
 Then, to ensure the narrative consistency of the simplified text, we synthesize auxiliary \textit{narrative prompts} combining key phrases derived from the syntactical analyses of the original text.
  Our model achieves results significantly 
  better than the seq2seq baseline on an 
  English medical corpus, yielding 3\%$\sim$4\% absolute improvements in terms of lexical similarity, and providing a further 1.1\% improvement of SARI score when combined with the baseline.
  We also highlight shortcomings of existing evaluation methods, and introduce 
  new metrics that take into account both lexical and high-level semantic similarity.
  A human evaluation conducted on a random sample of the test 
  set further establishes the effectiveness of the proposed approach. Codes and models are released here: \url{https://github.com/LuJunru/NapSS}.
\end{abstract}

\section{Introduction}

\begin{figure}[t]
  \centering
  \includegraphics[width=\linewidth]{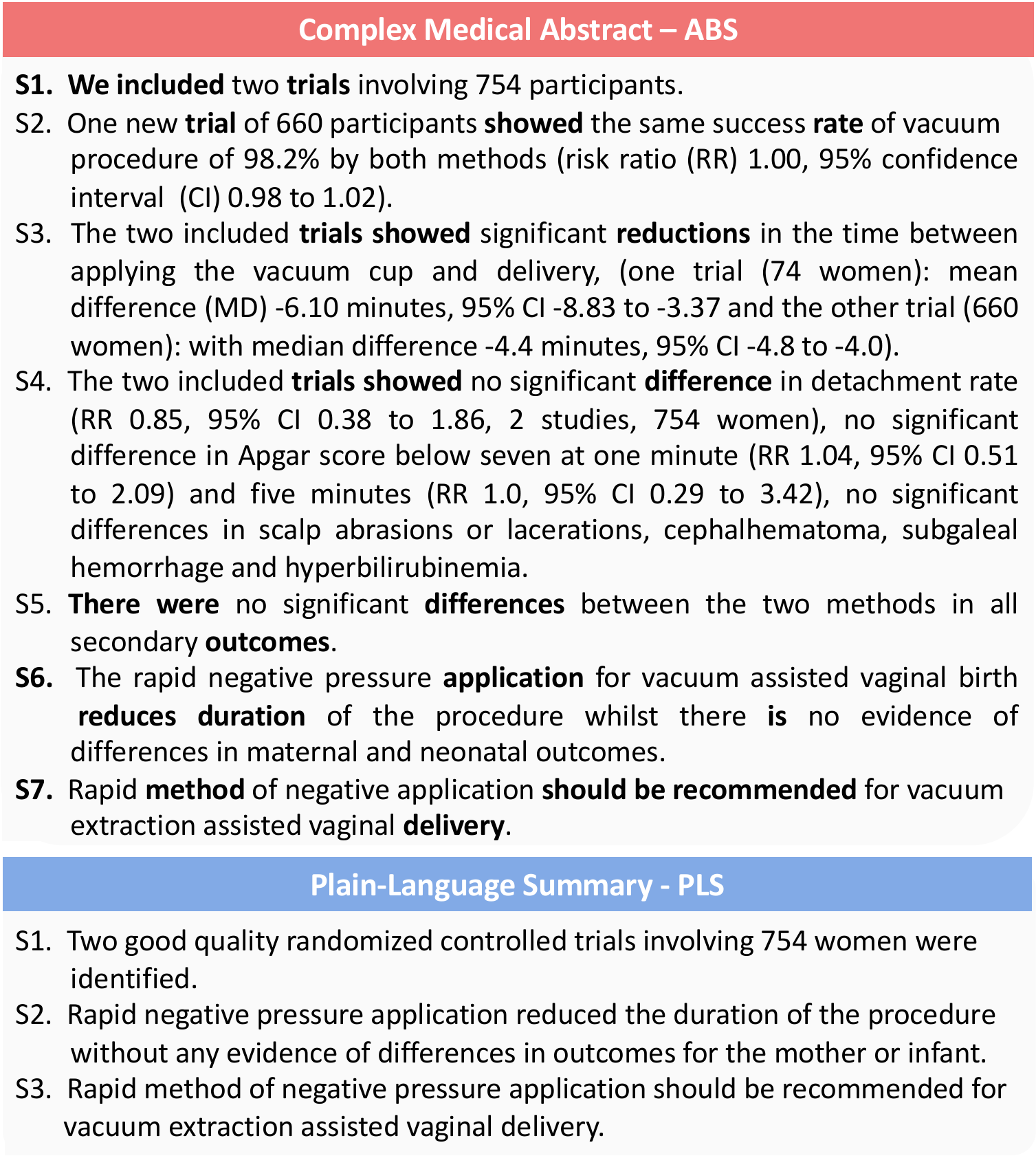}
  \caption{A typical sample of Medical Text Simplification task. The abstract and plain-language summary are split into sentences for easy inspection. Key phrases in each sentence, and marks of chosen sentences in reference summary are in bold.}
  \label{fig:sample}
\end{figure}

The medical literature is vast, and continues to expand quickly. Most patients (laypeople), however, are unable to access this information because it is written for specialists and so dense and laden with jargon. As the recent ‘infodemic’ has shown, access to reliable and comprehensible information about citizens’ health is a fundamental need: for example, a European Health Literacy Survey (HLS-EU) reports that "at least 1 in 10 (12\%) respondents show insufficient health literacy and almost 1 in 2 (47\%) has insufficient or problematic health literacy" \cite{sorensen15}.
Automated text simplification methods offer a potential means to address this issue, and make evidence available to a wide audience as it is published. However, performing paragraph-level simplification of medical texts is a challenging NLP task.

Online medical libraries such as Cochrane library,\footnote{\url{https://www.cochranelibrary.com/}} provide synopses of the medical literature across diverse topics, and manually-written plain language summaries. 
We are 
interested in developing accurate automated medical text simplification systems upon those libraries to help timely popularization of medical information to lay audience. 
We show a typical example of a technical abstract and associated simplified summary from a recently introduced paragraph-level medical simplification corpus \cite{2021Paragraph} in Figure \ref{fig:sample}. The sample consists of a technical abstracts (ABS) written for experts, and an manually authored Plain-Language Summaries (PLS) of the same publication collected from the Cochrane website. 
The dataset only provide raw abstract-PLS pairs. 
For easy inspection, we further add sentence splitting and highlight key phrases. 

As this example illustrates, a text simplification system needs to first have an overview of the key details reported in the abstract (e.g., that the review synthesizes `\emph{two trials}') and must also infer that there `\emph{were no significant differences}' when `\emph{rapid negative pressure application}' was applied to all participants, and thus that the `\emph{rapid method should be recommended}'. This entails an overall understanding of the key concepts to simplify, while preserving a consistent narrative flow. 
Built upon this general framing, the system should identify that the most representing sentences in the abstract are sentences 1, 6, 5 and 7. 
The key challenges here for a model include: (i) identifying the most important content to simplify within the synopsis; (ii) preserving the original narrative flow from a linguistic and medical point of view; (iii) synthesising the findings in a simple and consistent language.

To address these challenges, we propose a \emph{summarize-then-simplify} two-stage framework \texttt{NapSS}---\textbf{Na}rrative \textbf{P}rompting and \textbf{S}entence-matching \textbf{S}ummarization---for paragraph-level medical text simplification. 
The \textit{narrative prompt} is designed to promote the factual and logical consistency between abstracts (ABSs) and PLSs, while the \textit{simplification-oriented summarizer} identifies and preserves the relevant content to convey and simplify. 

In the first stage, we construct intermediate summaries via sentence matching between the abstract and the PLS sentences based on their Jaccard Distance. This preliminary set of summaries is used to fine-tune a simplification-oriented summarizer which at inference time identifies and extracts the most relevant content to be simplified from the technical abstracts. This extractive summarizer is simplification-aware in that the reference summary is built with PLS ground truth. 

In the second stage of simplification, the intermediate summary is concatenated to a narrative prompt generated by synthesising the main concepts, entities, or events mentioned in text resulting from the syntactic analysis of the PLSs.
The prepared input is passed to a seq2seq model (e.g., BART \cite{lewis2019bart}) to produce a plain-language output. 


Our contributions can be summarized as follows: 
\begin{itemize}
    \item We introduce \texttt{NapSS}, a two-stage \emph{summarize-then-simplify} approach for paragraph-level medical text simplification, leveraging extractive summarization and narrative prompting.
    \item We design a \textit{simplification-aware summarizer} and a narrative prompt mechanism. The former is based on a Pre-trained Language Model (PLM) fine-tuned for extractive summarization on an intermediate set of summaries built via sentence matching between the technical and simplified text. The latter synthesises key concepts from the medical text by syntactic dependency parsing analyses, promoting the overall consistency with the narrative flow.
    \item We conduct a thorough experimental assessment on the Cochrane dataset for paragraph-level medical simplification, evaluating the different features of the generated text (i.e., simplicity and semantic consistency) using several automatic metrics, and the model generalization on sentence-level simplification. Additionally, to mitigate the limitations of the automatic metrics, we designed and conducted a human evaluation assessment, involving ``layperson'' readers and medical specialists. The results demonstrated the state-of-the-art performance on quality and consistency of the simplified text.   
\end{itemize}

\section{Related work}
We review three lines of work relevant to this effort: text simplification, extractive summarization, and prompting.

\subsection{Text Simplification}

Work on text simplification has mainly focused on sentence-level simplification, using the Wikipedia-Simple Wikipedia aligned corpus \cite{zhu2010monolingual,woodsend2011learning} and the Newsela simplification corpus \cite{xu2015problems}. 
There has been less work on document-level simplification, perhaps owing to a lack of resources \cite{sun2021document,alva2019cross}. 

The medical domain stands to benefit considerably from automated simplification: The medical literature is vast and technical, and there is a need to make this accessible to non-specialists  \cite{Kickbusch2013HealthL}.
Some research uses those medical documents and deploys various simplification methods based on lexical and syntactic simplification \cite{damay2006simtext,kandula2010semantic,llanos2016managing}. The recent release of the Cochrane dataset provided a new parallel corpus of technical and lay overview of published medical evidence \cite{2021Paragraph}. 

\begin{figure*}[t]
  \centering
  \includegraphics[width=1.0\linewidth]{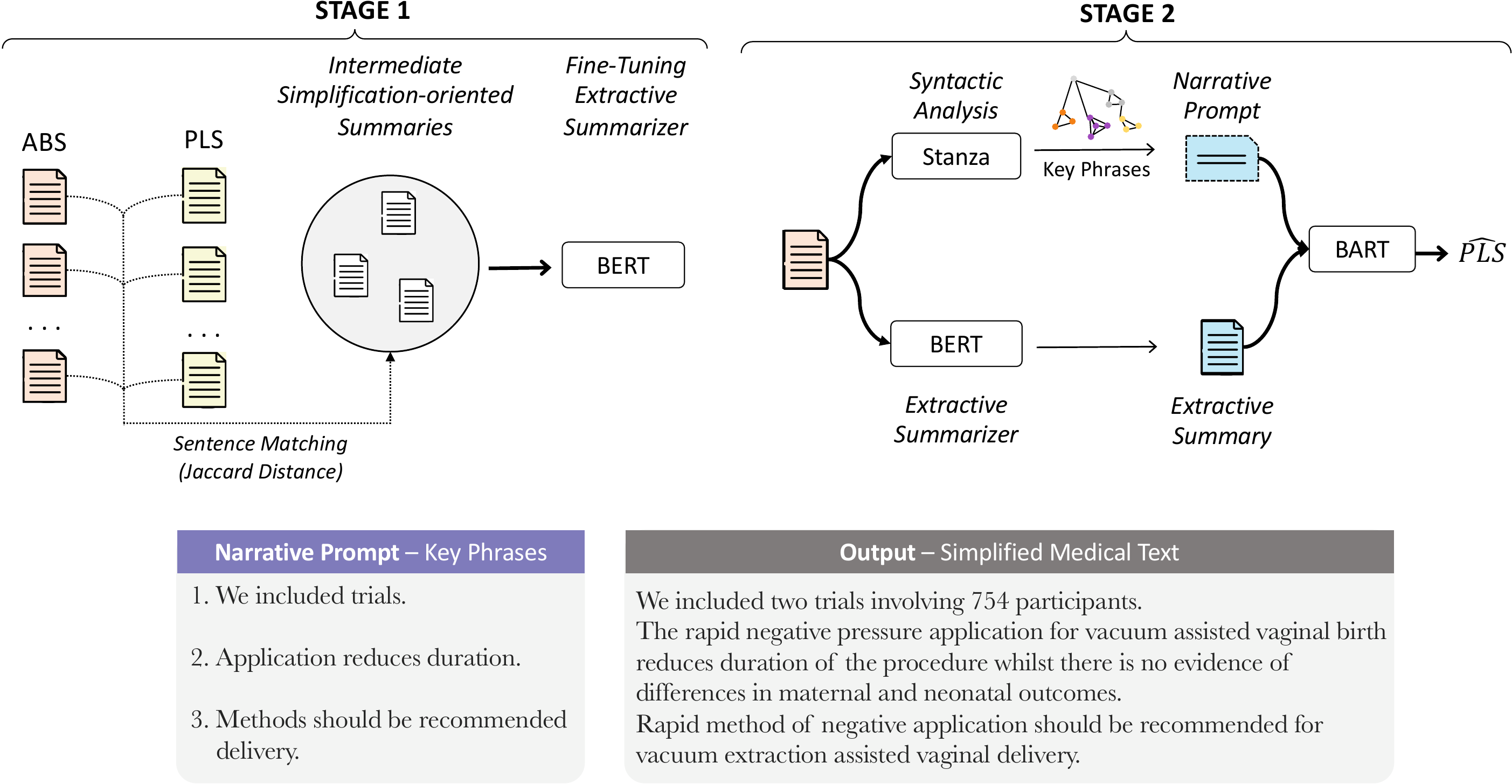}
  \caption{Overview of the two-stage pipeline in the \texttt{NapSS} model. 
  In the first stage, 
  we perform sentence ``labelling'' using Jaccard Distances \cite{jaccard1912distribution} over abstract (ABS) sentences in reference to PLS sentences, generating a set of intermediate summaries. A binary BERT-based \cite{devlin2018bert} classifier is fine-tuned over these summaries
  and used, at test time, to generate an extractive summary $\bm{x'}$. During the second stage (right side),
  we perform syntactic dependency parsing over the PLS sentences to extract key phrases $\bm{k}$. These are concatenated to form a narrative prompt and combined with the extractive ABS summary to serve as input of the simplification module for the generation of plain-language outputs $\bm{\hat{PLS}}$. In the bottom part, we reported an example of narrative prompt and simplified text generated by NapSS on the ABS introduced in Figure \ref{fig:sample}.} 
  \label{fig:model}
\end{figure*}

\subsection{Extractive Summarization}
Extractive summarization aims to select the most important words, sentences, or phrases from input texts and combine them into a summary. Many approaches have been proposed: ranking and selecting sentences based on their graph overlap \cite{mihalcea2004textrank}, deriving the relevance of the sentences within the text using WordNet \cite{6779492}, extracting information by named entity recognition \cite{maddela2022entsum}, and using continuous vector representations to perform semantic matching and sentence selection \cite{liu2019text, narayan2018ranking,gui19,lu2020chime,disen21}. 

There are some works that focus on extractive summarization of biomedical texts \cite{mishra2014text,sun22phee}. 
These have either aimed to provide a summary via graph-based methods or via sequence extraction to present key information in structured (tabular) form \cite{gulden2019extractive,aramaki2009text2table}.
In this work we follow a standard \emph{sentence matching} extractive summarization method   \cite{goldstein1999summarizing,zhong2020extractive} and fine-tune a pre-trained language model to perform sentence classification. 
We use extractive summaries as an intermediate step. 

\subsection{Prompting}

Recent work has shown that language models can be \emph{prompted} to perform tasks without supervision (i.e., ``zero-shot'') \cite{radford2018improving,NEURIPS2020_1457c0d6}. 
Prompts have been shown to work across a wide range of NLP tasks, e.g., sentiment classification, ``reading comprehension'', and ``commonsense reasoning'' \cite{seoh2021open,petroni2019language,boost21,DBLP:journals/corr/abs-1911-12543,lu22event,zhu22dis,wei2022chain}. 
Recent work has shown that prompt-based methods can be used even with smaller language models \cite{DBLP:journals/corr/abs-2001-07676,DBLP:journals/corr/abs-2012-15723}. 
In this work we focus on a novel use of prompts: Assisting generation of simplified text.

\section{Methods}
We first define the Paragraph-level Text Simplification task, introducing the relevant notations, and then present the \texttt{NapSS} model.

\subsection{Task Formulation}
In many cases, text simplification can be viewed as a generative task with additional constraints regarding the simplicity of the generated text.  
Analogously to text summarization, paragraph-level text simplification can be formulated as follows: for a given \textit{complex} paragraph with \textit{M} sentences, $\bm{x}=\{\{x^1_{1},x^1_{2},\cdots,x^1_{N_{x^1}}\}\cdots\{x^M_{1},x^M_{2},\cdots,x^M_{N_{x^M}}\}\}$, the aim is to generate  a \textit{plain-language summary} (PLS) $\bm{\hat{y}}=\{\hat{y}_{1},\hat{y}_{2},\cdots,\hat{y}_{N_s}\}$, summarizing and simplifying the original paragraph, with $N_{x^m}$ 
denoting the length of the $m$-th sentence $\bm{x^m}$. 


\subsection{NapSS}
We now describe \texttt{NapSS}, a text simplification approach based on a \emph{summarize-then-simplify} two-stage pipeline with the aims of (i) identifying the relevant content to simplify while (ii) ensuring that the original narrative flow is preserved. 
First, we generate a preliminary summary by using a \textit{simplification-oriented BERT summarizer}, an extractive model fine-tuned beforehand to identify the most relevant content to attend and simplify (§\ref{subsec:sum}).
These preliminary summaries are then combined with a \textit{narrative prompt}, a synthetic set of key phrases describing the main concepts, entities, or events discussed in the original text and derived from its syntactic analysis (§\ref{subsec:np}).
The overall working flow of our proposed \texttt{NapSS} model is illustrated in Figure \ref{fig:model}.
We next provide the details of each of these modules.

\subsubsection{Sentence-matching Summarization}
\label{subsec:sum}
The idea behind the summarization stage is to identify the most important content within a given technical abstract (with respect to target simplifications). 
We automatically construct an intermediate ``reference'' summary dataset using the simplification training set with which to fit a simplification-oriented summarizer.
Specifically, we train the latter as a binary sentence classifier, which provides a simple extractive summarization approach. 

\begin{algorithm}[t]
  \begin{algorithmic}[1]
    \STATE Input require: abstract sentence sets \{$\bm{x^m_{1 \sim M}}$\}, 
    \STATE PLS sentence sets \{$\bm{y^q_{1 \sim Q}}$\}
    \STATE Initilization: Empty positive sentence set $x_{pos}$
    \FOR{PLS sentence $\bm{y^q} \in \{\bm{y^q_{1 \sim Q}\}}$}
      \STATE Initilization: Minimum Jaccard Distance $\text{Dist}_{q} \leftarrow 10.0$, 
      \STATE corresponding sentence index $\text{Ind}_{q} \leftarrow 0$
      \FOR{abstract sentence $\bm{x^m} \in \{\bm{x^m_{1 \sim M}}\}$}
        \STATE ${\text{Dist}}_{qm} = {\text{\tt JaccardDistance}}(\bm{y^q}, \bm{x^m})$
        \IF{$\text{Dist}_{qm} < \text{Dist}_{q}$}
        \STATE $\text{Dist}_{q} \gets \text{Dist}_{qm}$
        \STATE $\text{Ind}_{q} \gets m$
        \ENDIF
      \ENDFOR
      \IF{$\bm{x^{\text{Ind}_{q}}} \notin x_{pos}$}
      \STATE add $\bm{x^{\text{Ind}_{q}}}$ in $x_{\text{pos}}$
      \ENDIF
    \ENDFOR
    \STATE Negative sentence set $x_{\text{neg}} = \{\bm{x^m_{1 \sim M}}\} - x_{\text{pos}}$
  \end{algorithmic}
  \caption{Build reference summary dataset}
  \label{alg:building}
\end{algorithm}

Algorithm \ref{alg:building} details the process of building this pseudo reference summary dataset. 
The input to the algorithm are the sets of sentences from the technical abstract (ABS) and 
the corresponding simplified text (PLS).
For each PLS sentence, 
we calculate the Jaccard Distance to every ABS sentence, 
and select the one with the lowest score. 
The set of selected ABS sentences constitute an intermediate extractive summary of the technical abstract.
The complexity of Algorithm \ref{alg:building} is $O(N_x \cdot N_y \cdot D)$, where $D$ denotes the size of entire corpus.

Based on the intermediate summary dataset, we fine-tune a BERT model to perform binary classification over sentences. 
At inference time, the resultant  trained \textit{simplification-oriented summarizer} is used to select sentences from the technical abstract which will be simplified.
These are concatenated and then passed to a BART model \cite{lewis2019bart} along with the narrative prompt. 

As an example, the bottom left of Figure \ref{fig:model} shows 3 PLS sentences guiding the automatic labelling (0/1) of 7 ABS sentences. 
The intermediate extracted summary $\bm{x'}$ derived via Jaccard matching is used at training time, while 
at inference time we extract this using the trained model. 

\subsubsection{Narrative Prompting}
\label{subsec:np}
Intuitively, the simplification-oriented summarizer should identify the most important content in ABS which should be simplified. 
However, the similarity matching with which we train the sentence classifier may be noisy and miss relevant information constituting the narrative flow, resulting in errors that lead to omissions in outputs. 
Therefore, in our \texttt{NapSS} model, we incorporate another simple mechanism, \emph{narrative prompting}, to encourage factual consistency between the input and output. 

Inspired by recent work on \emph{chain-of-thought} ``reasoning'' \cite{wei2022chain}, we assume a logical narrative chain can be explicitly constructed with key phrases extracted via syntactic dependency parsing, and then used as a prompt. 
Specifically, we use a light natural language processing tool Stanza\footnote{\url{https://stanfordnlp.github.io/stanza/}} for dependency parsing on every abstract sentence to extract key phrases. 
Algorithm \ref{alg:keys} details the algorithmic process of our narrative prompting. The algorithm takes abstract sentences as input, runs a dependency parse on each, collects the root token and its closest child tokens to form key phrases in natural linguistic orders, and assembles these as the narrative prompt. Let $k^m$ denotes the key phrase of sentence $\bm{x^m}$, the narrative prompt $k^M$ equals to $[k^1$\texttt{</s>}$k^2$\texttt{</s>}$\cdots$\texttt{</s>}$k^m]$, in which ``\texttt{</s>}'' is a special separation token. The complexity of this building algorithm is $O(N_x \cdot D)$. As shown in Figure \ref{fig:model}, key tokens are shown with bold fonts in every abstract sentences.

\begin{algorithm}[t]
  \begin{algorithmic}[1]
    \STATE Input require: abstract sentence sets \{$\bm{x^m_{1 \sim M}}$\}
    \STATE Initilization: Empty key phrases queue $x_{que}$
    \FOR{abstract sentence $\bm{x^m} \in \{\bm{x^m_{1 \sim M}}\}$}
      \STATE DTree = DependencyParsing($\bm{x^m}$)
      \STATE $\bm{x^m_{root}}$ = DTree.Root()
      \STATE $\bm{x^m_{root_l}}$, $\bm{x^m_{root_r}}$ = DTree.Children($\bm{x^m_{root}}$)
      \STATE $k^m$ = $\bm{x^m_{root_l}}$ $\bm{x^m_{root}}$ $\bm{x^m_{root_r}}$
      \STATE add $k^m$ in $x_{que}$
    \ENDFOR
    \STATE Prompt $k^M = k^1$</s>$k^2$</s>$\cdots$</s>$k^m$
  \end{algorithmic}
  \caption{Build narrative prompt}
  \label{alg:keys}
\end{algorithm}

\subsubsection{Text Simplification}
The resulting input of the second text simplification stage is composed by $[k^M$\texttt{</s>}$\bm{x'}]$, as depicted in the bottom right part of Figure \ref{fig:model}. \texttt{NapSS} adopts encoder-decoder PLM models as the backbone for generative text simplification. Let $L_{gen_{TS}}$ be the loss of the generative text simplification task:
\begin{equation}
  L_{gen_{TS}} = -\frac{1}{N_k + N_{x'}}\sum_{t=1}^{N_k + N_{x'}}y_{t}\log \hat{y}_{t}
\end{equation}
where $N_k$, $N_{x'}$ are the lengths of the narrative prompt $k^M$ and of the extractive summary $\bm{x'}$, respectively.

\section{Experimental Assessment}

\subsection{Experimental Setup}

\paragraph{Dataset} 
We build and evaluate \texttt{NapSS} on the first published paragraph-level medical text simplification dataset \cite{2021Paragraph}. 
The dataset is derived from the Cochrane library of systematic reviews and contains 4,459 parallel
pairs of technical (ABS) and simplified (PLS) medical abstracts curated by domain experts. 
The average length of abstract is around 300 to 700 tokens, while the average length of PLS is around 130 to 390 tokens \cite{2021Paragraph}. 
All abstract and PLS text are preprocessed to have a total token length lower than 1,024, which is a typical input upper bound of large PLM models. 
The dataset was split into 3,568 training, 411 development and 480 testing instances. 
To our knowledge, this is the only accessible paragraph-level text simplification dataset.

For the summarization model, the derived summary dataset contains 51,635 training, 5,856 development, and 7,009 testing sentences (constructed from the respective dataset splits). This dataset contains around 53\% positive sentences and 47\% negative sentences, which is relatively balanced, and consistent with the proportion of average amount of PLS sentences and average amount of paired abstract sentences. 
We describe hyperparameter selection in the Appendix Section \ref{app:hyper}.

\begin{table*}[t]
\resizebox{\textwidth}{!}{
  \begin{tabular}{l|cc|cccc|c|c|c}
    \toprule
    \quad & \multicolumn{2}{c|}{Readability} & \multicolumn{4}{c|}{Lexical Similarity} &  \multicolumn{1}{c|}{Simplification} & \multicolumn{1}{c|}{Semantic Similarity} & \multicolumn{1}{c}{Comprehensive}\\
    \toprule
    Models & FK & ARI & Rouge-1 & Rouge-2 & Rouge-L & BLEU & SARI & BertScore & BLEURT\\
    \midrule
    Vanilla BART & 10.89 & 14.32 & 46.79 & 19.23 & 43.55 & 11.5 & 38.72 & 23.94 & -0.194\\
    UL-BART \cite{2021Paragraph} & 11.97 & 13.73 & 38.00 & 14.00 & 36.00 & 39.0 & 40.00 & / & /\\
    UL-BART (\texttt{by us}) & 9.30 & 12.40 & 43.25 & 16.36 & 40.22 & 7.9 & 40.08 & 24.64 & -0.309\\
    \midrule
    \texttt{NapSS} (\texttt{our}) & 10.97 & 14.27 & \textbf{48.05} & 19.94 & 44.76 & \textbf{12.3} & 40.37 & \textbf{25.73} & \textbf{-0.155}\\
    \texttt{NapSS} BioBART & 10.98 & 14.24 & 47.66 & 19.77 & 44.39 & 11.9 & 40.21 & 25.61 & -0.166\\
    \midrule
    \texttt{NapSS} (\texttt{+UL}) & \textbf{8.67} & \textbf{11.80} & 45.39 & 16.77 & 42.53 & 9.1 & \textbf{41.12} & 23.13 & -0.219\\
    \texttt{NapSS} (\texttt{-Prompt}) & 9.86 & 13.06 & 45.62 & \textbf{20.01} & \textbf{44.83} & 12.1 & 39.68 & 25.57 & -0.158\\
    \texttt{NapSS} (\texttt{-Summary}) & 10.62 & 13.99 & 46.91 & 19.51 & 44.18 & 11.8 & 39.62 & 25.29 & -0.167\\
    \bottomrule
  \end{tabular}}
    \caption{Overall results on the testing set. UL BART is the previous SOTA, and we report results from our re-implementation of this. The inconsistency between \cite{2021Paragraph} and our re-implementation is due to the inavailability of evaluation code. For \texttt{NapSS}, we provide 2 groups of results by changing backbone model of text simplification module. The robustness verification of proposed \texttt{NapSS} is provided in appendix \ref{app:robust}. We further provide fusion and ablation results based on BART version of \texttt{NapSS}. \texttt{NapSS} (-Prompt) refers to remove the narrative prompt, while \texttt{NapSS} (-Summary) is to replace the abstract summary with full abstract.}
  \label{tab:results}
\end{table*}

\paragraph{Evaluation Metrics}
For evaluation we largely adopt the metrics used in prior work on this task and dataset \cite{2021Paragraph}.
These can be placed into three groups: readability metrics, lexical similarity metrics, and simplification metrics. The readability metrics include the Flesch–Kincaid grade level score (FK) \cite{kincaid1975derivation} and the automated readability index
(ARI) \cite{senter1967automated}. Lexical similarity metrics are widely adopted to evaluate text generation, including ROUGE-1, ROUGE-2, ROUGE-L \cite{lin2004rouge} and BLEU \cite{papineni2002bleu}. The simplification metrics include SARI \cite{xu2016optimizing}, which is an editing-base metric especially designed for text simplification task. In our setting, SARI would reward the \textit{generation} of words occurring only in the paired PLSs, and avoidance of ABS words not occurring in the corresponding PLS. 

Simple automated metrics fail to capture semantic agreement between outputs and references. 
We therefore consider two additional metrics: BertScore \cite{zhang2019bertscore} and BLEURT \cite{sellam2020bleurt}. 
BertScore was originally designed to evaluate semantic similarity via BERT \cite{devlin2018bert} embeddings. \citet{simplicity_da} and \citet{devaraj2022evaluating} recently assessed and verified its effectiveness on the text simplification task. 
BLEURT is a metric finetuned on both lexical BLEU metric and semantic BertScore metric. Along with the automatic assessment, we also conduct a manual (human) evaluation of the simplicity, fluency and factuality whose evaluation criteria are detailed Section §\ref{sec:human_eval}.

Prior work did not publicly provide code to perform evaluations beyond computing ROUGE.\footnote{\url{https://github.com/AshOlogn/Paragraph-level-Simplification-of-Medical-Texts}} 
Therefore, we mainly compare 
results according to our re-implementation of evaluation metrics. 

\paragraph{Baseline} ``Vanilla'' BART is a pretrained encoder-decoder architecture, based on transformers, whose auto-regressive decoder made it a suitable a strong baseline for text generation. In ours setting, we adopted a a specific checkpoint version\footnote{\url{https://huggingface.co/facebook/bart-large-xsum}} additionally fine-tuned on the XSUM dataset \cite{narayan2018don, devaraj2021paragraph}, providing higher performance on text summarization.
The only other 
model developed for paragraph-level medical text simplification is \textit{UL-BART} \cite{devaraj2022evaluating}, is also based on BART but integrates an auxiliary ``unlikelihood'' (UL) penalty 
to demote generation of technical jargon, which improved the readability and simplicity of outputs compared to the base BART model.

\subsection{Results}
\subsubsection{Automatic Metrics} We report quantitative results in Table \ref{tab:results} comparing the main models and the ablation studies. 
We notice that UL-BART can generate text which is more readable (lower FK and ARI) and simpler (higher SARI) than the ``Vanilla'' BART.
However, the model struggles to maintain lexical and semantic similarity (lower ROUGE, BLUE, and higher BLEURT) to the human references, perhaps because omitting jargon terms as the modified objective degrades coherence. 

By contrast, \texttt{NapSS} improves lexical similarity by 3\% to 4\% in terms of ROUGE and BLEU scores while maintaining a comparable SARI score.
\texttt{NapSS} additionally improves the semantic similarity between the model outputs and the human references at the cost of a slightly higher FK and ARI scores, demonstrating an higher semantic consistency while simplifying the medical text. 
For the sake of completeness, we also tested whether replacing the ``Vailla'' BART backbone with a specialised medical PLM, such as BioBART \cite{yuan2022biobart}, would lead to better performance. Surprisingly, the replacement 
did not lead to any significant change in any of the adopted metrics.

We further explored the integration of the auxiliary ``unlikelihood'' (UL) loss in \texttt{NapSS} (+UL), aiming at increasing the degree of simplification while preserving semantic consistency.
The resulting model yielded further state-of-the-art performance on the overall text simplicity with an increase of \char`~0.8\%  in readability and 1.1\% in SARI score. 
\texttt{NapSS} (-Prompt) and (-Summary) refer to two ablation models. The first one removes the narrative prompt, leading to improved readability but decreased simplification (lower SARI). The second one show that the full abstract is necessary for improving the lexical similarity.

We report and discuss in Appendix \ref{app:summarizer} the binary classification performance of the extractive summarization module used in stage one.

\begin{figure*}[t]
  \centering
  \includegraphics[width=1.0\linewidth]{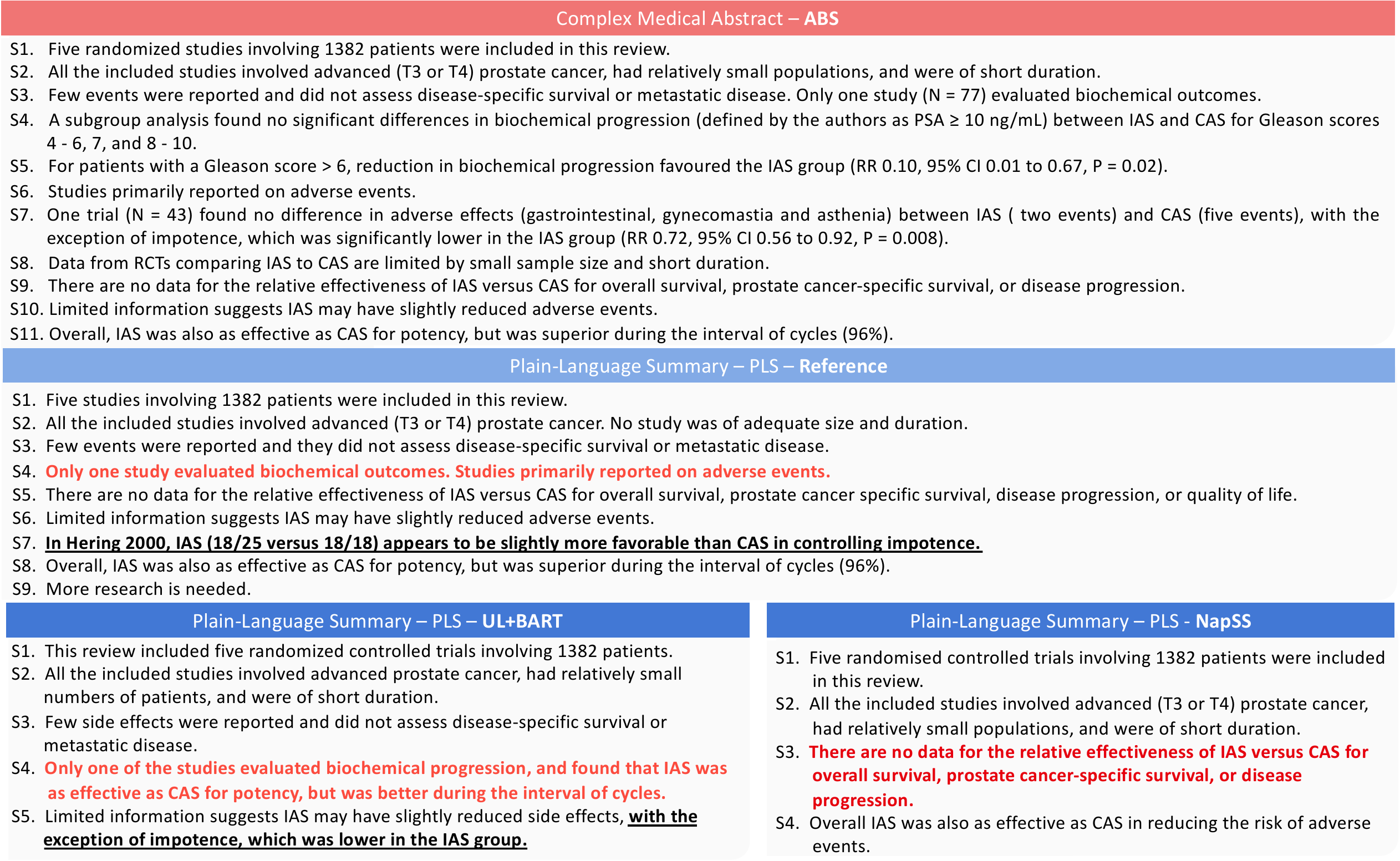}
  \caption{Case study and error analysis on a typical example from the testing set. Smeared sentences illustrate factual improvement by \texttt{NapSS}, while underlined parts reveal information omission of our model outputs.}
  \label{fig:cases}
  \vspace{-10pt}
\end{figure*}

\subsubsection{Out-of-Domain Evaluation}
To evaluate the generalization ability of \texttt{NapSS}, we evaluate the model on
a \emph{different} medical text simplification dataset: TICO-19 \cite{shardlow2022simple}. 
Unlike the Cochrane dataset, this is designed for sentence-level simplification and contains over 6k parallel technical and simplified sentences related to COVID-19.

Table \ref{tab:TICO19} reports results. The ``Vanilla'' BART and UL-BART have the best performance on readability 
while \texttt{NapSS} yields over \char`~2\% improvement in terms of simplicity. Integrating \texttt{NapSS} with the ``unlikelihood'' (UL) penalty (\texttt{NapSS} (\texttt{+UL}))  achieves around \char`~1-3\% boost on lexical and semantic evaluation. The overall results highlight  that our approach can preserve a high level of semantic consistency for simplification at the sentence level, yet with slightly reduced readability. 

\begin{table}[t]
\resizebox{\columnwidth}{!}{
  \begin{tabular}{lccccc}
    \toprule
    Models & FK & ARI & BLEU & SARI & BLEURT\\
    \midrule
    Vanilla BART & 4.91 & \textbf{6.83} & 9.71 & 43.47 & -0.663\\
    UL-BART (\texttt{by us}) & \textbf{4.76} & 7.61 & 8.75 & 40.83 & -0.654\\
    \texttt{NapSS} (\texttt{our}) & 6.32 & 7.99 & 10.1 & \textbf{45.78} & -0.648\\
    \texttt{NapSS} (\texttt{+UL}) & 5.49 & 8.25 & \textbf{12.8} & 44.46 & \textbf{-0.553}\\
    \bottomrule
  \end{tabular}}
  \caption{Zero-shot inference results. All above models are only fine-tuned on the Cochrane dataset \shortcite{2021Paragraph}, then run zero-shot inference on the TICO-19 test set.}
  \label{tab:TICO19}
\end{table}

\subsubsection{Human Evaluation}
\label{sec:human_eval}
We designed and conducted a manual evaluation of the outputs generated by the simplification models to provide additional insights into \textit{fluency} and \textit{factuality}; the latter is especially difficult to assess with existing automatic metrics. 

\paragraph{Evaluation Procedure}
We randomly sampled 100 unsimplified instances (ABSs) from the test set and paired each with simplified outputs 
generated by two models, one from UL-BART \cite{2021Paragraph} and one from the proposed {\tt NapSS}. 
Each simplified text was assessed by three different annotators. 
We hired 6 annotators to participate in this evaluation, who are postdoctoral researchers and PhD students in computer science. 
Each was assigned 100 instances; this took nearly 8 hours to complete. 
Additionally, we hired two expert annotators who have professional background in the medical domain to obtain a reliable evaluation on the factual consistency between the complex and the simplified text. Annotators were paid \$19 per hour.
To ensure that annotators shared a common understanding of our evaluation criteria, we held a tutorial session with detailed instructions and provided 20 instances 
as a trial run. We then resolved any annotation inconsistencies afterwards.

\paragraph{Evaluation Criteria}  
We followed a previous approach to ask annotators to give numerical scores for each instance \cite{simplicity_da}. 
Considering the requirement for the simplification tasks and text styles characterizing medical documents \cite{devaraj2022evaluating}, we separated numerical scores into three aspects: simplicity, fluency and factuality. 
Annotators can select a numerical rating (from 0, 1, and 2) for each aspect. 
Appendix \ref{app:annotat} provides details for each category. 

\paragraph{Results}
\begin{table}[t]
 \centering
 \resizebox{\columnwidth}{!}{
 \begin{tabular}{lccccc}
 \toprule
 Models & Simplicity & Fluency & Factuality & (Experts) & Overall\\ \midrule
 UL BART (by us)  & \textbf{1.43} & 1.53 & 1.17 & 0.99 & 4.13 \\
 \texttt{NapSS} & 1.12 & \textbf{1.54} & \textbf{1.66} & \textbf{1.28}&\textbf{4.32} \\
 \bottomrule
 \end{tabular}}
 \caption{Human evaluation result by each category.}
 \label{tab:human_eval}
\end{table}

In Table \ref{tab:human_eval}, we present average annotator scores assigned to all aspects. 
Our model achieves higher overall and average scores on Fluency and Factuality, respectively. 
UL-BART model got higher score on Simplicity because this model sometimes generates too simple outputs. Simplicity from our evaluation schema only focuses on evaluating the length of the text and the vocabulary. It does not involve the evaluation of the content. Therefore, if the generated text only contains a conclusion from the paragraph, our evaluator would give a higher score on Simplicity. On the contrary, the fluency and factuality aspects focus on evaluation at the context and semantic level, where our model got a higher score in the assessment. As Factuality is an aspect that the evaluation is subject to evaluators' background knowledge, therefore we selected those instance been given three different scores from basic evaluators to create an experts set. We can see experts' evaluation also shows the same trend. We believe the narrative prompt benefits this improvement. Our model tends to produce a reasonable reduction in the context while keeping the majority of critical points. It is also useful for the model to calibrate grammar and plausibility with prompts. Combined with narrative prompt, \texttt{NapSS} generates simplification more consistent with the original text than the UL-BART. 
We can observe the better performance on human evaluation results also correlated with the improvement in semantic and comprehensive metrics, which proves the necessity of semantic level simplification evaluation.



\subsubsection{Case Study and Error Analysis}
We present a case study and error analysis based on the examples reported in Figure \ref{fig:cases}.\footnote{Better with colors} The \texttt{Abstract} (ABS) mentions the the analysis of 5 studies on the effects of the continuous (CAS) or intermittent (IAS) androgen suppression therapy on advanced prostate cancer. The UL-BART model generated a slightly longer simplified text than \texttt{NapSS}. Specifically, sentence 4 from the UL-BART output mixed and linked the biochemical progression assessment with the IAS and CAS side-effect for potency.
In contrast, sentence 3 generated by \texttt{NapSS} is more relevant to 
the findings of all studies considered. 

On the other hand, the last sentence 5 from the UL output reported a meaningful finding in consistence with reference sentence 7 from the PLS and reference sentence 7 from the ABS. \texttt{NapSS} instead omitted this information, probably because the related PLS sentences were not considered sufficiently relevant by the model.

\section{Conclusions}
We proposed a \emph{summarize-then-simplify} two-stage model---\texttt{NapSS}---for paragraph-level medical text simplification. 
The first component is a ``simplification-oriented'' summarizer, which we trained over a heuristically derived set of ``psuedo'' references derived via sentence matching.
At inference time, the summarizer extracts the most relevant content to be simplified.
This is combined with an additional ``narrative prompt'' intended to promote consistency, and then passed to an encoder-decoder model to produce the simplified text.
Experiments on a paragraph-level medical text simplification showed that, under several automatic metrics and human evaluation (involving ``laypeople'' and medical specialists), this method realized significant improvements with respect to both simplification quality and consistency. 
\section*{Limitations}
Our study is primarly based on the Cochrane paragraph-level medical text simplification dataset \cite{2021Paragraph}.
While this dataset provides richer and more elaborated text  than previous sentence-level medical datasets, such as TICO-19 \cite{shardlow2022simple}, it is worth noting that experimental documents tend to share 
a common pattern whose structure consists of: (i) discussing statistics about the clinical trials considered, (ii) list the experimental assessments, (iii) summarize the conclusions of the related findings. 

Despite the already significant difficulty of the task, a limited variety of documents would inevitably introduce linguistic bias, hindering the model generalization and our current ability to conduct thorough assessment of the methodologies.

Moreover, although we made effort to examine the factuality aspect with expert annotators, we acknowledge that factuality is a subjective aspect and existing methods may not be sufficient to verify.


\section*{Ethics Statement}
This work is based on publicly available medical datasets \cite{2021Paragraph,shardlow2022simple}. As stated by the authors of datasets, no personal identification information were released. 
Current language technologies generally---and automated simplification models such as the one proposed in this work---still introduce ``hallucinations'' and factual inaccuracies into outputs; at present we would therefore recommend against deploying fully automated generative models for medical texts.

\section*{Acknowledgment}
This work was supported in part by the UK Engineering and Physical Sciences Research Council (EP/T017112/1, EP/V048597/1, EP/X019063/1), and the National Science Foundation (NSF) grant 1750978. YH is supported by a Turing AI Fellowship funded by the UK Research and Innovation (EP/V020579/1). This work was conducted on the UKRI/EPSRC HPC platform, Avon, hosted in the University of Warwick’s Scientific Computing Group.
BCW was supported in this work by the National Institutes of Health (NIH), grant R01-LM012086. GP, JL, and JL, were supported by the National AI Strategy award (Warwick/ATI): `\textit{METU: An Inclusive AI-Powered Framework Making Text Easier to Understand}'.

\bibliography{custom}
\bibliographystyle{acl_natbib}

\appendix

\begin{table*}[t]
\resizebox{\textwidth}{!}{
  \begin{tabular}{l|cc|cccc|c|c|c}
    \toprule
    \quad & \multicolumn{2}{c|}{Readability} & \multicolumn{4}{c|}{Lexical Similarity} &  \multicolumn{1}{c|}{Simplification} & \multicolumn{1}{c|}{Semantic Similarity} & \multicolumn{1}{c}{Comprehensive}\\
    \toprule
    Models & FK & ARI & Rouge-1 & Rouge-2 & Rouge-L & BLEU & SARI & BertScore & BLEURT\\
    \midrule
    \texttt{NapSS} (seed=42) & 10.97 & 14.27 & 48.05 & 19.94 & 44.76 & 12.3 & \textbf{40.37} & \textbf{25.73} & -0.155\\
    \texttt{NapSS} (seed=123) & 10.89 & 14.17 & \textbf{48.38} & \textbf{20.24} & \textbf{45.11} & \textbf{12.5} & 40.36 & 25.67 & -0.149\\
    \texttt{NapSS} (seed=2023) & \textbf{10.85} & \textbf{14.09} & 48.29 & 20.09 & 45.02 & 12.4 & 40.31 & 25.60 & \textbf{-0.148}\\
    \bottomrule
  \end{tabular}}
    \caption{Robustness checking of our \texttt{NapSS}.}
  \label{tab:robust}
\end{table*}

\section{Experimental Setup}
\subsection{Hyperparameters}
\label{app:hyper}

For the summarization stage, we adopt NLTK\footnote{\url{https://www.nltk.org/}} for the building of reference summary dataset, and fine-tune a distilbert-base-uncased-finetuned-sst-2-english\footnote{\url{https://huggingface.co/distilbert-base-uncased-finetuned-sst-2-english}} PLM as the classifier. The chosen PLM is a distilbert-base-uncased\cite{Sanh2019DistilBERTAD} checkpoint additionally fine-tuned on SST-2 dataset\cite{socher2013recursive}, which is a sentiment binary classification corpus. The hidden size of the checkpoint is 768 and the corresponding vocabulary size is 30,522. The random seed is 42. The batch size is set to 16 and the accumulation steps is set to 1 on 2 quadro\_rtx\_6000 GPUs. The optimizer is BertAdam\footnote{\url{https://github.com/google-research/bert/blob/master/optimization.py}} with $\beta1=0.9, \beta2=0.999$, and $\epsilon$=1e-6. The weight of decay is 0.01. The learning rate is 2e-5 without warmup. It takes 0.5 hour in total to fine-tune the checkpoint on the training set, and predict over development and testing sets.

And for the simplification stage, except for possible replacement of backbone encoder-decoder PLM, we adopt exact same settings with the SOTA baseline \cite{2021Paragraph}, including training strategy and sampling method during the predictive generation. It takes less than 20 mins to fine-tune the PLM, while requires 2 hours to generate simplified text over entire testing set on same GPUs.

\subsection{Annotation Schema}
\label{app:annotat}

To overcome the aforementioned limitations on evaluation metrics, we followed a previous approach to ask our annotators give numerical scores for each instances \cite{simplicity_da}. Considering the requirement on simplification task and feature of text in medical domains \cite{devaraj2022evaluating}, we designed our numerical scores into three aspects: Simplicity, Fluency and Factuality. Annotator can select one numerical score under each aspect, which include three options 0,1 and 2. Higher score stands for annotator consider the paragraph level performance under that aspect is excellent, vice versa. In here, we provide detail explanation of each aspect. 

Simplicity aspect considers how simple that text is to read. This category assess the generated text by annotator's impression of simplicity, in terms of length of the texts and use of vocabulary. A good simplified text is expected to omit unnecessary numerical descriptions and explain jargons that are hard to be understood by layman readers.

Fluency aspect considers the how fluent the text is. That is, to assess the simplified text by annotator's impression on connectivity and fluency. A good simplified paragraph should consider the fluency among sentences, such as use of conjunction words or adversative words for sentences. This category also includes the evaluation on overall grammar correctness of each sentences, and penalty on duplicate sentences generated by the model.

Factuality considers how consistent is the simplified text with the original text. This category requires annotators to assess the generated text by compare the facts that mentioned from the original text and those included in the generated text. A good simplified text should includes all the important information appears in the original text. Any paraphrase on the simplified text that lead to different meaning and against the original texts, or any omits on important information should consider to give penalize under this category. 

\section{Robustness of \texttt{NapSS}}
\label{app:robust}

We finetune our \texttt{NapSS} model with another two random seeds 123 and 2023. The results of three experiments in \ref{tab:robust} share high similarity, confirming the robustness of our proposed pipeline.

\section{Summarizer Results}
\label{app:summarizer}
We fine-tuned two different bert-based classifiers, the aforementioned distillbert one, and another BioLinkBERT-base\footnote{\url{https://huggingface.co/michiyasunaga/BioLinkBERT-base}}, which is a bert-based model pretrained on PubMed abstracts concerning citation links\cite{yasunaga2022linkbert}. Although the BioLink backbone was pretrained on medical corpus, the general Distillbert fine-tuned on similar binary classification dataset performed better.

\begin{table}[h]
\resizebox{\columnwidth}{!}{
  \begin{tabular}{lcc}
    \toprule
    Models & Accuracy & F1\\
    \midrule
    BioLinkBERT-base & 61.91 & 67.04\\
    Distilbert-base-uncased-finetuned-sst-2-english & 62.50 & 68.91\\
    \bottomrule
  \end{tabular}}
  \caption{Performance on the constructed testing set.}
  \label{tab:summary}
\end{table}

\end{document}